\documentclass[runningheads]{llncs}
\usepackage{graphicx}
\usepackage{cite}
\usepackage{amsmath,amssymb,amsfonts}
\usepackage{algorithmic}
\usepackage{subfigure}
\usepackage{textcomp}
\usepackage{xcolor}
\usepackage{multirow}
\usepackage{float}
\usepackage{hyperref}
\usepackage[hyphenbreaks]{breakurl}
\begin{document}
\title{Subjective evaluation of traditional and learning-based image coding methods}
\titlerunning{Subjective evaluation of image coding methods}
%
\author{Zhigao Fang\inst{1} \and
    Jiaqi Zhang\inst{1} \and
    Lu Yu\inst{1} \and Yin Zhao\inst{2}}

\authorrunning{Z. Fang et al.}
%

\institute{Zhejiang Provincial Key Laboratory of Information Processing, Communication and Networking, Hangzhou, China\\
    \email{\{zhigao.fang, jiaqi.zhang, yul\}@zju.edu.cn}\\
    \and
    Huawei Technologies Co., Ltd., Hangzhou, China\\
    \email{yin.zhao@huawei.com}}
\maketitle              
\begin{abstract}
    We conduct a subjective experiment to compare the performance of traditional image coding methods and learning-based image coding methods. HEVC and VVC, the state-of-the-art traditional coding methods, are used as the representative traditional methods. The learning-based methods used contain not only CNN-based methods, but also a GAN-based method, all of which are advanced or typical. Single Stimuli (SS), which is also called Absolute Category Rating (ACR), is adopted as the methodology of the experiment to obtain perceptual quality of images. Additionally, we utilize some typical and frequently used objective quality metrics to evaluate the coding methods in the experiment as comparison. The experiment shows that CNN-based and GAN-based methods can perform better than traditional methods in low bit-rates. In high bit-rates, however, it is hard to verify whether CNN-based methods are superior to traditional methods. Because the GAN method doesn’t provide models with high target bit-rates, we can’t exactly tell the performance of the GAN method in high bit-rates. Furthermore, some popular objective quality metrics have not shown the ability well to measure quality of images generated by learning-based coding methods, especially the GAN-based one.

\end{abstract}
\section{Introduction}
In recent years, with the quick development of the Internet, image coding methods are utilized more widely because without coding, the images on the Internet can consume huge storage. Image coding methods can be mainly divided into two categories, one compresses images using traditional methods (e.g. transform, quantification, entropy coding) and the other is the learning-based methods, e.g. using neural network\cite{review}. Recently, the learning-based image coding methods like CNN based and Generative Adversarial Network (GAN) based methods significantly improve the compression efficiency. These methods leverage highly non-linear models, leading to quite different distortions in the decompressed images compared to those blocky and blur from traditional methods. \cite{RDP} shows that in mean square error in a decoded image, small distortion does not always mean good visual quality. Thus, it is insufficient to evaluate these learning methods only using some typical objective metrics (e.g. PSNR). The proper way to present the superiority of coding models is offering some results of subjective study. However, few works provide subjective result in their paper. As a result, we don't know whether their proposed models outperform these prior coding methods in visual quality.

We notice that Ascenso et al.\cite{ascenso2020learning} and Valenzise et al.\cite{QualityAss} have done some similar works on evaluating learning-based image coding methods. \cite{QualityAss} compared a CNN method\cite{balle2017end} and a RNN method\cite{toderici2017rnn} with JPEG2000 and BPG. \cite{ascenso2020learning} also used the aforementioned RNN model and a CNN model\cite{balle2018variational} as the representive learning-based methods and chose JPEG, HEVC, WebP as anchors. Both of them adopted Double Stimulus Impairment Scale (DSIS) as their subjective test protocol. They have shown that JPEG is no more advanced and HEVC intra coding (BPG) does not always outperform the learning-based image coding methods.

Prior works have done well on measuring the performance of learning-based coding methods with the DSIS protocol. However, these works actually measure the degradation of impairing images compared to reference images\cite{itu-t}, not the perceptual quality (i.e. the image quality measured only with distorted images)\cite{perceptualquality}. In our opinion, the perceptual quality is an important attribute of an image. Thus, this paper aims to find out the perceptual quality of the images compressed by learning-based coding methods especially CNN and GAN methods compared to those coded by traditional coding methods. Two state-of-the-art traditional coding methods and three advanced learning-based coding methods including two CNN methods and a GAN method are utilized in the subjective experiment. Single Stimuli, also called Absolute Category Rating (ACR), is adopted as the test protocol. The experiment shows that the well-trained GAN and CNN based image coding methods can provide better human visual perception in low bit-rates than traditional image coding methods, but in high bit-rates, for most of the images, the performance of CNN methods are similar to the traditional methods.

The rest of the paper is structured as follow: Section 2 describes the details of the subjective experiment, including the image set we used, the metrics used to assess images and the organization of the experiment. Section 3 analyses the result the experiment and Section 4 draws the conclusion.

\section{Design of the Subjective Evaluation}
The experiment aims to find out the performance of the learning-based coding methods when they are applied to the practical applications as in practice, reference images are usually not available. This section describes the details of our experiment. Firstly, we lay out the images, then describe the coding methods we select. Finally we describe the procedure of the subjective test.
\subsection{Image set}\label{imageset}
\begin{figure}[t]
    \centerline{\includegraphics[width=0.8\textwidth]{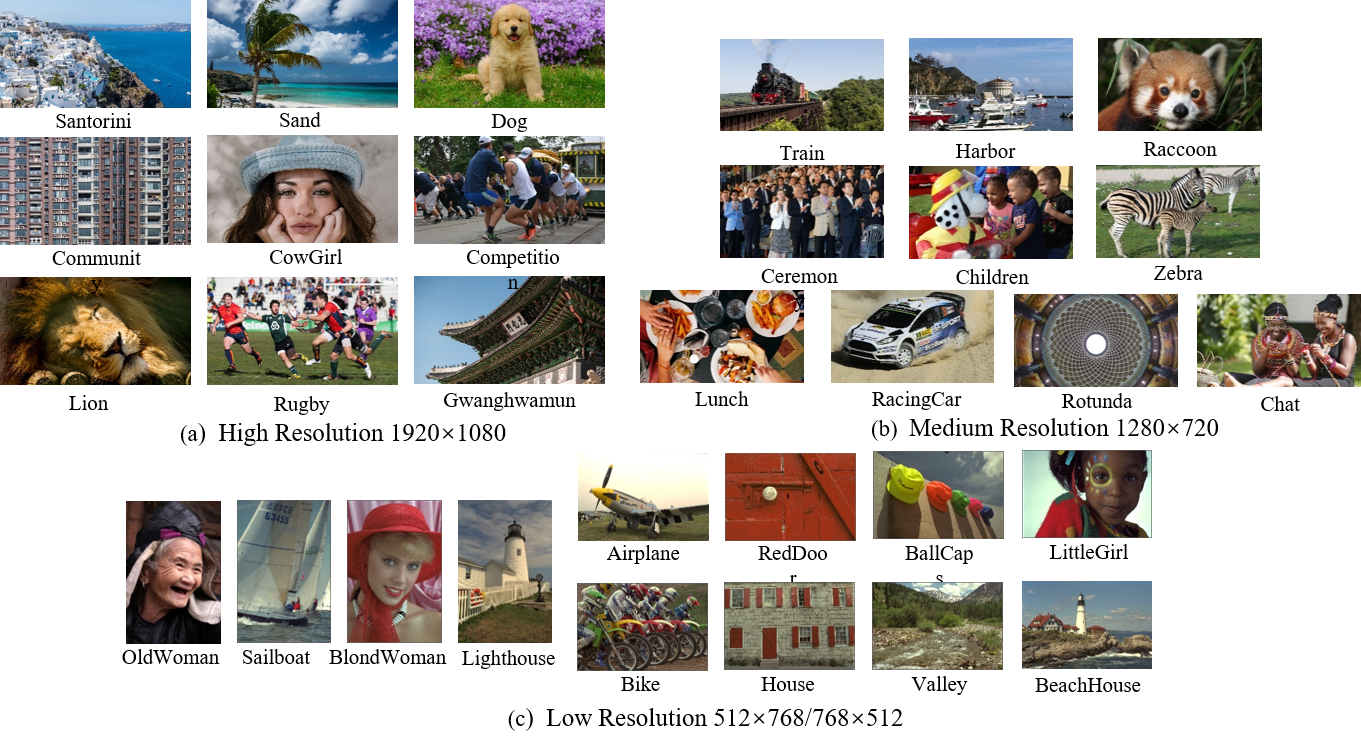}}
    \caption{The images we select.}
    \label{fig_dataset}
\end{figure}

At present, the commonly used high quality datasets in the learning-based coding methods are Kodak PhotoCD\cite{kodak}, DIV2K\cite{div2k}, CLIC\cite{clic2020}, Cityscapes\cite{cityspace}, etc. Some CNN and RNN based end to end image coding methods have achieved high MS-SSIM score on the CLIC dataset\footnote{http://challenge.compression.cc/leaderboard/lowrate/test/}. Agustsson et al. proposed a GAN-based extreme learned compression model\cite{gc} that performs well on the Kodak PhotoCD (768×512) and Cityscapes (down scale to 1024×512). Mentzer et al. proposed a GAN-based high-fidelity compression model\cite{mentzer2020high} that achieved about 0.95 MS-SSIM score on the DIV2K and CLIC2020.
We select images from Kodak PhotoCD dataset, DIV2K dataset and JPEG AI Test Dataset\footnote{Available on https://jpegai.github.io/test\_images/}\cite{jpegai}, which is aimed at evaluating the performance of the training models. Then these images are cropped and compressed by some image coding methods. After getting the processed images, a small scale experiment is conducted to remove the images in which the distortion caused by coding methods is hard to see. Eventually, we obtain 31 images, including 9 high resolution images (1920×1080), 10 medium resolution images (1280×720) and 12 low resolution images (512×768 and 768×512). All images selected can be seen in Fig.~\ref{fig_dataset}.

\subsection{Coding methods}
This section introduces the coding methods used in the experiment. When selecting the coding methods, to ensure that the experiment is significant, we prefer those that are advanced or typical.

Learning-based image coding techniques mainly contain CNN, RNN, GAN, etc. CNN and GAN typically use the auto-encoder architecture as their backbone and use some entropy coding methods to encode the latent representation of the model as the compression result\cite{review}. Many works have shown learning-based coding methods have powerful capability to improve the compression efficiency. Ballé et al. proposed a CNN model using a scale hyperprior that achieved similar performance with BPG in PSNR and MS-SSIM\cite{balle2018variational}. Lee et al. proposed a context-adaptive entropy model that achieved better performance than BPG\cite{Lee2019Context}. Cheng et al. proposed a discretized Gaussian mixture likelihoods based entropy model that achieved similar performance to VTM 5.2 in PSNR\cite{Cheng}. Agustsson et al. proposed a GAN based compression method, operating images at extremely low bit rates\cite{gc}. Later Mentzer et al. proposed a High-Fidelity GAN model called HIFIC\cite{mentzer2020high} that can reconstruct images more similar to the input images than \cite{gc} at extremely low bit-rate for high resolution images. Considering the performance as well as whether the code is available, we select \cite{Cheng} as a representive CNN method and \cite{mentzer2020high} as a representive GAN method in the experiment. Because \cite{balle2018variational} is instructive to the subsequent work on CNN-based image coding methods, we choose it as another CNN coding method. We have five pre-trained models of \cite{Cheng}, of which the target bit-rate are 0.1, 0.2, 0.3, 0.7, 0.8. The HIFIC has three pre-trained models available with three target bit-rates (i.e. 0.14bpp, 0.3bpp, 0.45bpp). We retrain the HIFIC and get a new model with a smaller target bit-rate (i.e. 0.06bpp). The models of \cite{balle2018variational} we used are provided by \cite{begaint2020compressai}, which has several quality levels. Both of the CNN methods are optimized by MSE and the GAN method is optimized by LPIPS and MSE. We finally choose five levels (i.e. 1, 2, 3, 5, 7) to compress images because images compressed with these levels can have similar bit-rates with aforementioned two learning-based methods.

Currently, the most widely used traditional image lossy coding method is JPEG, but \cite{ascenso2020learning} and \cite{QualityAss} have shown that JPEG and its improved version, JPEG 2000, are no more advanced than HEVC. As the inheritor of HEVC, VVC is indisputably better than HEVC\cite{Cheng}. Therefore, We decide to use HEVC and VVC as two traditional image coding methods in the experiment. The software of HEVC we used is BPG\cite{bpg}, an integration of HEVC intra coding and the software of VVC is VTM. Because the learning-based methods we use only have finite models and the target quality of different methods is different, we have to try a number of Quality Points (QPs) to ensure that the images compressed by traditional methods are similar to those by aforementioned learning-based coding methods. Eventually, the five QPs of BPG utilized are 48, 44, 38 32, 26 and the QPs of VTM are 40, 39, 35, 30, 22. The version of BPG we use is 0.9.8 and VTM is 10.0.

In summary, we utilize five coding methods to compress our images, which are BPG, VTM, two CNNs and one GAN. The GAN method has four target quality and others have five target quality or Quality Points. The total number of images compressed is 744.
\begin{figure}[tp]
    \centerline{\includegraphics[width=0.8\textwidth]{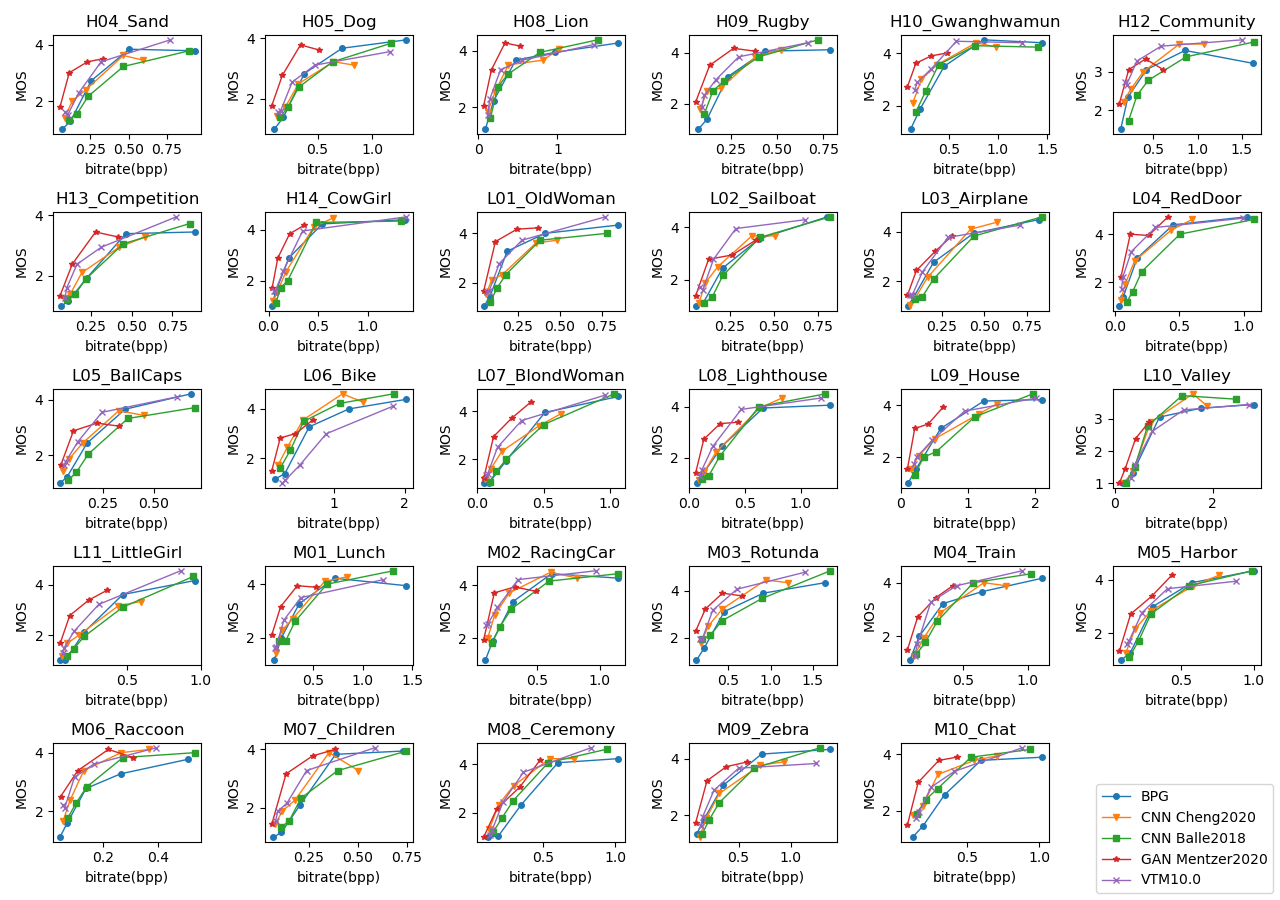}}
    \caption{The Rate-MOS curves for all images in the experiment.}
    \label{R-MOS curve}
\end{figure}
\subsection{Subjective test procedure}
The subjective test can adopt Single Stimulus (SS) method or Double Stimulus Impairment Scale (DSIS) method as its test protocol\cite{itu-t}. DSIS method presents both an original image and a corresponding decompressed image to the observers and then asks observers to rate the impairment of the decompressed image compared to the original image. The result of this method can reflect the fidelity of the decompressed image with respect to the source image. Single Stimulus method shows an image to observers and asks them to evaluate the quality of the image with their experience, of which the result shows the visual quality of an image without reference. It is similar to the scene that people watch images on the Internet. Because the experiment aims to find out which image coding methods can provide decompressed images with better visual quality in the scene that only decompressed images are available, Single Stimulus method is used as the test methodology in our experiment. We use five-point scale (1 to 5) to present the image quality, labeled as bad, poor, fair, good and excellent\cite{itu-t}.

A 46 inch Hyundai monitor with a $1920\times1080$ resolution is used to conduct the experiment. Because the images in our image set have three different size and scaling may change the quality of images, the images are presented on the screen with their original size. If an image can’t fill the whole screen, the rest of the screen is filled with gray. To confirm the observers can see distortions in images clearly, we keep the watching distance three times the images' height\cite{itu-t}. In other words, the distance between observers and the monitor is variable depending on the size of an image. For the images with the resolution 512x768, we set the distance to the three times images' width so that when watching the images with 768x512 and 512x768 resolution, observers can have the same angle of viewing. Before the experiment, we pick out one high resolution image and one low resolution image with their relevant coding images, the number of which are 26, to pre-train
observers so that they can know the relation between image quality and five-point scores. Besides, we put 29 uncompressed images into the experiment to conform whether the observers give consistent ratings. During the experiment, we show one image for 6 seconds and give observers 5 seconds to decide the score of the image quality\cite{itu-t}. The experiment is divided into three sessions. In each session, observers take a short break for every 20 minutes. Observers are required to complete the three sessions at three different time and at the beginning of each session, observers are pre-trained by the aforementioned sample images. There are totally 19 volunteers participating in our experiment and each volunteer score 722 images.
\begin{figure}[tp]
    \centering
    \includegraphics[width=0.8\textwidth]{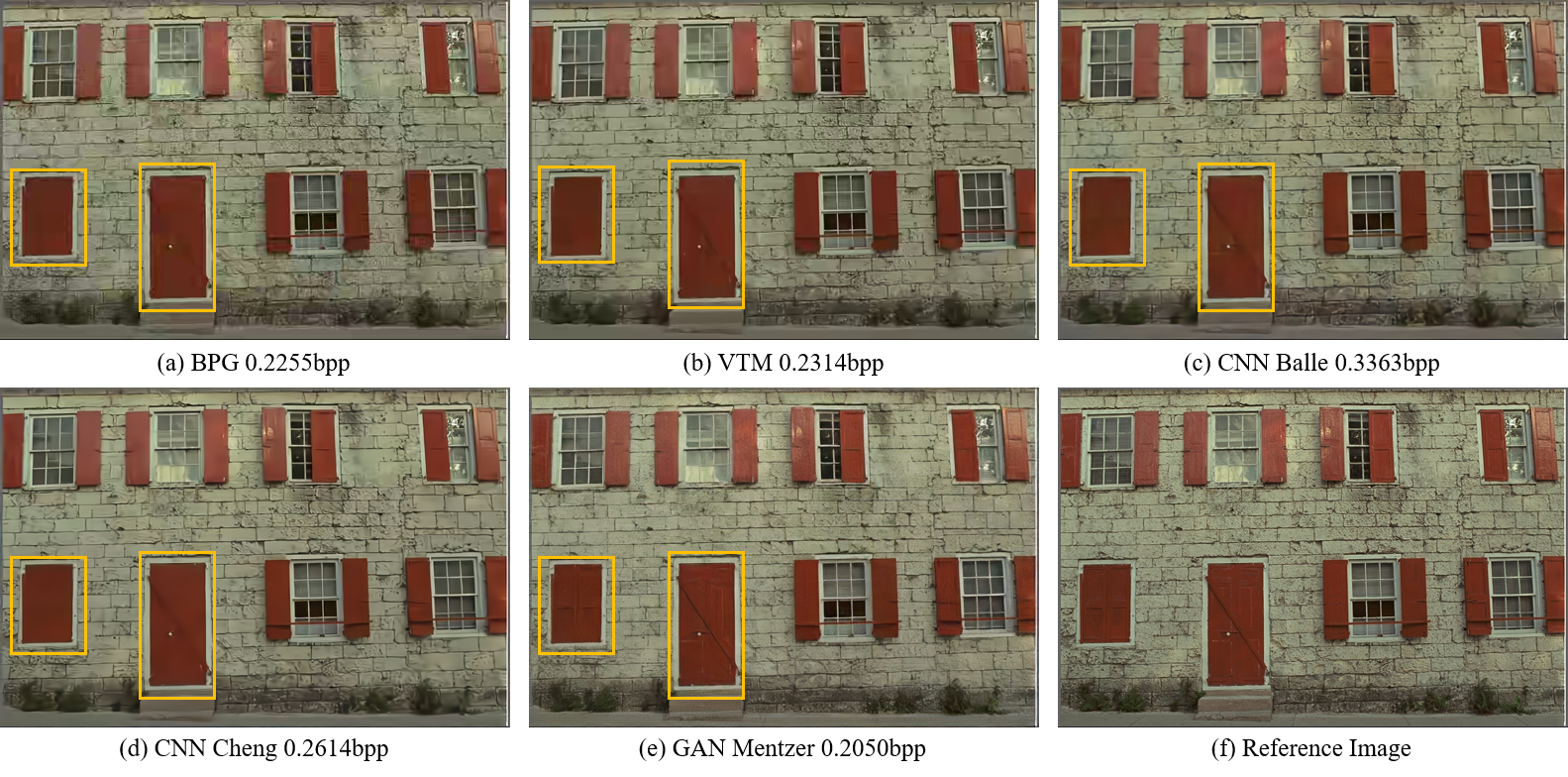}
    \caption{The "L09" compressed by BPG, VTM, CNN Balle, CNN Cheng, GAN and reference image from left to right, top to down. The yellow bounding boxes mark out some obvious difference among five methods in low bit-rates}
    \label{fig_L09}
\end{figure}
\subsection{Subjective Score Processing}
Before calculating the mean opinion scores, screening of the observers is performed to detect the unqualified observers according to ITU-R BT.500-14\cite{itu-r}. 18 volunteers' scores are accepted after the procedure. Following the screening, the Mean Opinion Score of each image is calculated with:
\begin{equation}
    MOS_j=\frac{1}{N}\sum_{i=1}^Nu_{ij},\label{eq3}
\end{equation}
where $u_{ij}$ indicates the score of the $j$-th image given by $i$-th volunteer.

\section{Experimental results}
\begin{figure}[t]
    \centering
    \includegraphics[width=0.8\textwidth]{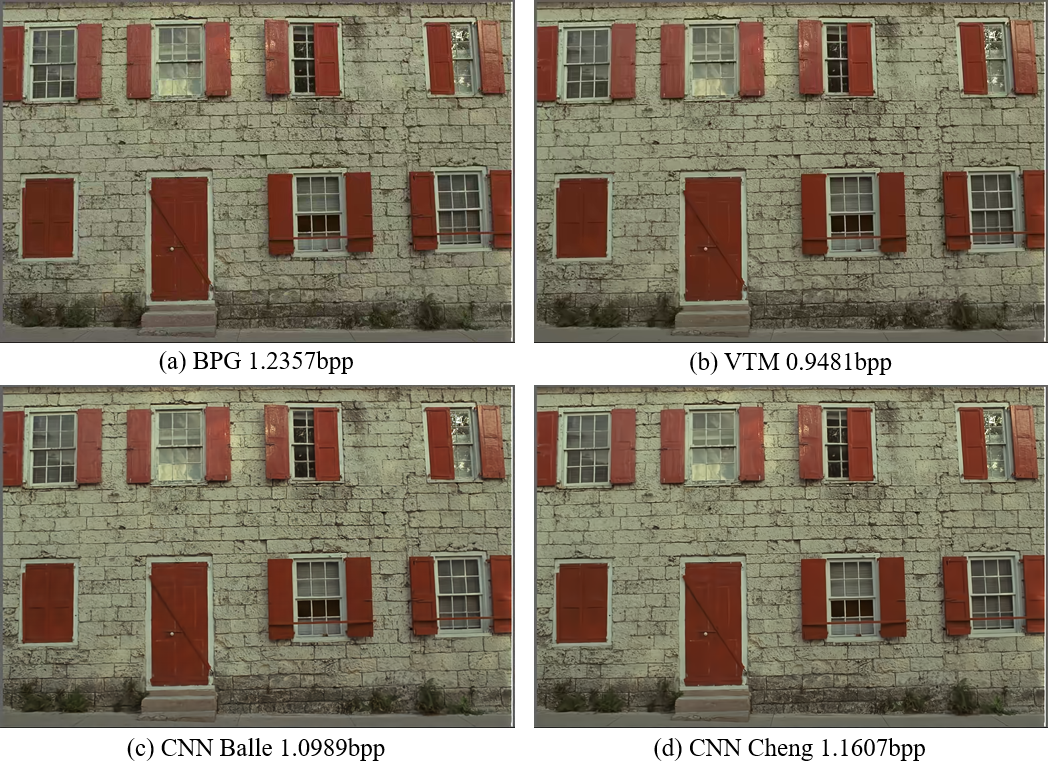}
    \caption{The "L09" compressed by BPG, VTM, CNN Balle and CNN Cheng from left to right, top to down in high bit-rates}
    \label{fig_L09_high}
\end{figure}

\begin{figure}[tp]
    \centering
    \includegraphics[width=0.8\textwidth]{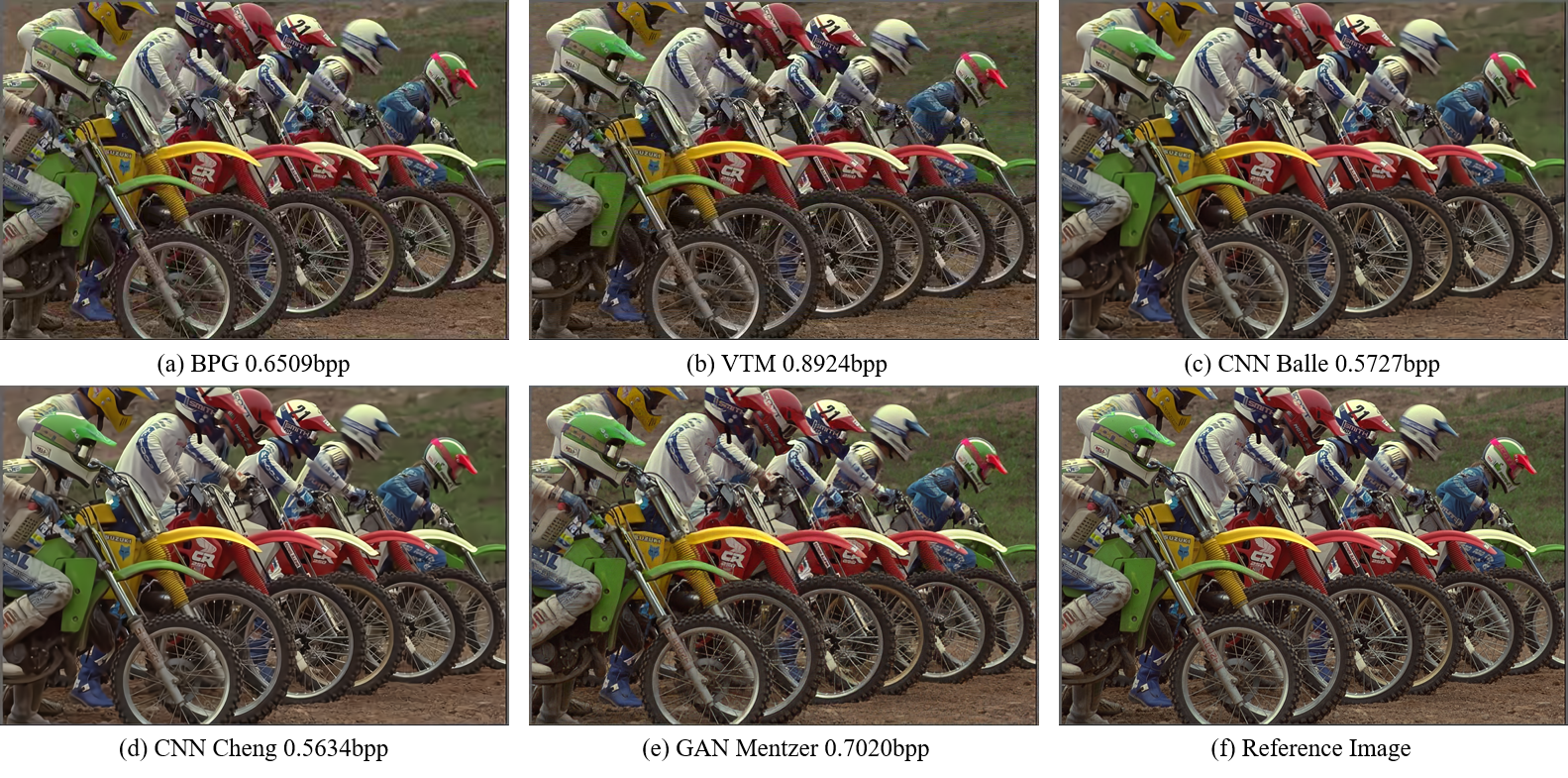}
    \caption{The "L06" compressed by BPG, VTM, CNN Balle, CNN Cheng, GAN and reference image from left to right, top to down}
    \label{fig_L06}
\end{figure}

The Rate-MOS curves of all the images utilized in the experiment are showed in Fig.~\ref{R-MOS curve}. From the curves we can see that for most images, the GAN method selected in the experiment outperforms other coding methods in low bit-rates, e.g. below 0.25 bpp. For instance, Fig.~\ref{fig_L09} presents the impairing images of image "L09" compressed by five methods. It is obvious that the details of the red door and window in yellow rectangles are impaired heavily except for the image generated by GAN. This result shows that the GAN method can better recover some textures of the coding images in decoding in low bit-rates, which makes it perform better than other aforementioned coding methods. When the bit-rates of images become higher (e.g. above 1 bpp), the result is different. The MOS of most images in high bit-rates is close to each other. From the Fig.~\ref{fig_L09_high}, it is found that all of the four coding methods recover the image "L09" well. The details of the door and windows can be seen clearly and even the texture of the wall is similar to the reference image, which is shown in Fig.~\ref{fig_L09}. Thus, it is hard to distinguish which coding method is best. Because the GAN method doesn't provide the models for high target quality, its performance on high bit-rates is not clear enough. Additionally, it can be seen that in image "L06", VTM performs worst in the five coding methods and BPG is also worse than the learning models, which is out of our expectation. The "L06" shows in Fig.~\ref{fig_L06}. It can be seen that the texture of people and motorcycles in five images is not so bad, but the backgrounds on the top-right of five images are all impairing to some degree compared to the reference image. BPG and VTM lead to slightly blocky distortion. What's different is that the distortion caused by VTM is some small horizontal blocks, which makes the green color spread and makes the image look worse than BPG's. Two CNN methods and the GAN method cause blur distortion. The MOS result of "L06" shows that subjects seems to be more tolerant to blur distortion than blocky distortion for this image.

To show the performance difference between different coding methods more visually. We figured out the BD-MOS-rate\cite{bjontegaard2001calculation} of the four coding methods using BPG as the anchor, which shows in Table~\ref{tab1}. The result shows that the CNN model proposed by Cheng, VTM and the GAN model are all better than BPG. As the quality range of the GAN model is smaller than the VTM's, we can't draw the conclusion that the GAN coding method is superior to VTM.
\begin{table}[t]
    \caption{BD-MOS-rate of the experiment result. BPG are used as the baseline.}
    \label{tab1}
    \centering
    \begin{tabular}{|c|r|r|r|r|}
        \hline
        \multirow{2}{*}{\textbf{Image}} & \multicolumn{1}{c|}{\textbf{CNN}}       & \multicolumn{1}{c|}{\textbf{CNN}}         & \multicolumn{1}{c|}{\multirow{2}{*}{\textbf{VTM}}} & \multicolumn{1}{c|}{\textbf{GAN}}         \\
                                        & \multicolumn{1}{c|}{\textbf{Cheng2020}} & \multicolumn{1}{c|}{\textbf{ball\'e2018}} & \multicolumn{1}{c|}{}                              & \multicolumn{1}{c|}{\textbf{Mentzer2020}} \\ \hline
        L01                             & 16.04\%                                 & 39.98\%                                   & -7.86\%                                            & -51.67\%                                  \\ \hline
        L02                             & -26.51\%                                & 7.56\%                                    & 20.19\%                                            & -40.25\%                                  \\ \hline
        L03                             & 3.09\%                                  & 27.72\%                                   & -16.98\%                                           & -33.09\%                                  \\ \hline
        L04                             & -2.41\%                                 & 67.42\%                                   & -34.78\%                                           & -45.10\%                                  \\ \hline
        L05                             & -51.15\%                                & 32.72\%                                   & -28.63\%                                           & -54.82\%                                  \\ \hline
        L06                             & -25.22\%                                & -24.48\%                                  & -8.15\%                                            & -47.88\%                                  \\ \hline
        L07                             & -20.52\%                                & 6.00\%                                    & -34.11\%                                           & -59.35\%                                  \\ \hline
        L08                             & -0.96\%                                 & 12.90\%                                   & -25.60\%                                           & -58.79\%                                  \\ \hline
        L09                             & 8.44\%                                  & 36.48\%                                   & -0.50\%                                            & -57.16\%                                  \\ \hline
        L10                             & -1.86\%                                 & -16.09\%                                  & 2.29\%                                             & -24.40\%                                  \\ \hline
        L11                             & -17.58\%                                & 14.92\%                                   & -30.64\%                                           & -59.33\%                                  \\ \hline
        M01                             & -84.45\%                                & -86.88\%                                  & -87.38\%                                           & -97.93\%                                  \\ \hline
        M02                             & -6.42\%                                 & 24.82\%                                   & -25.10\%                                           & -50.65\%                                  \\ \hline
        M03                             & -40.34\%                                & 12.93\%                                   & -33.32\%                                           & -62.77\%                                  \\ \hline
        M04                             & -5.17\%                                 & 12.91\%                                   & -19.49\%                                           & -37.95\%                                  \\ \hline
        M05                             & -6.24\%                                 & 7.01\%                                    & -16.66\%                                           & -39.90\%                                  \\ \hline
        M06                             & -40.38\%                                & -19.07\%                                  & -100.00\%                                          & -58.46\%                                  \\ \hline
        M07                             & 8.15\%                                  & 13.05\%                                   & -26.94\%                                           & -55.45\%                                  \\ \hline
        M08                             & -39.88\%                                & -20.71\%                                  & -35.03\%                                           & -35.07\%                                  \\ \hline
        M09                             & 14.18\%                                 & 32.60\%                                   & -10.73\%                                           & -46.59\%                                  \\ \hline
        M10                             & -31.62\%                                & -25.43\%                                  & -27.18\%                                           & -59.93\%                                  \\ \hline
        H04                             & 54.84\%                                 & 27.60\%                                   & -31.66\%                                           & -60.24\%                                  \\ \hline
        H05                             & 78.75\%                                 & 14.05\%                                   & -7.46\%                                            & -72.46\%                                  \\ \hline
        H08                             & -3.11\%                                 & 1.50\%                                    & -17.77\%                                           & -64.78\%                                  \\ \hline
        H09                             & -10.30\%                                & -6.64\%                                   & -26.84\%                                           & -79.50\%                                  \\ \hline
        H10                             & -10.26\%                                & -3.84\%                                   & -23.19\%                                           & -64.95\%                                  \\ \hline
        H12                             & -25.93\%                                & 22.39\%                                   & -25.84\%                                           & -48.31\%                                  \\ \hline
        H13                             & -2.59\%                                 & 11.61\%                                   & -28.36\%                                           & -63.81\%                                  \\ \hline
        H14                             & 13.00\%                                 & 28.95\%                                   & -2.17\%                                            & -50.53\%                                  \\ \hline
        \textbf{Average}                & -8.84\%                                 & 8.69\%                                    & -24.48\%                                           & -54.52\%                                  \\ \hline
    \end{tabular}
\end{table}

As many works offer some objective quality assessment metrics to present their performance, we also evaluate the five coding methods by PSNR as well as the structural similarity assessments, MS-SSIM\cite{msssim}\cite{msssimcode}. Additionally, we use Netflix’s VMAF\cite{vmaf}, LPIPS\cite{lpips} and DISTS\cite{DISTS}. The PSNR is computed on the Y channel of the YCrCb space. Similarly, the MS-SSIM is also computed on the Y channel. VMAF is aimed at evaluating the perceptual video quality by fusing various quality assessment metrics. Follow the requirement of VMAF, we first transform images from RGB color space to YCrCb color space and then compute the VMAF scores. The VMAF version we use is 0.6.1 and the VDK is 1.5.1. LPIPS evaluates the image quality using perceptual similarity. Before inputting the reference image and reconstructed image to the LPIPS v0.1 API, we normalize images to $[-1, 1]$ as it requires. The images evaluated by DISTS are also need to be normalized to $[0, 1]$.

\begin{figure}[t]
    \centering
    \includegraphics[width=0.8\linewidth]{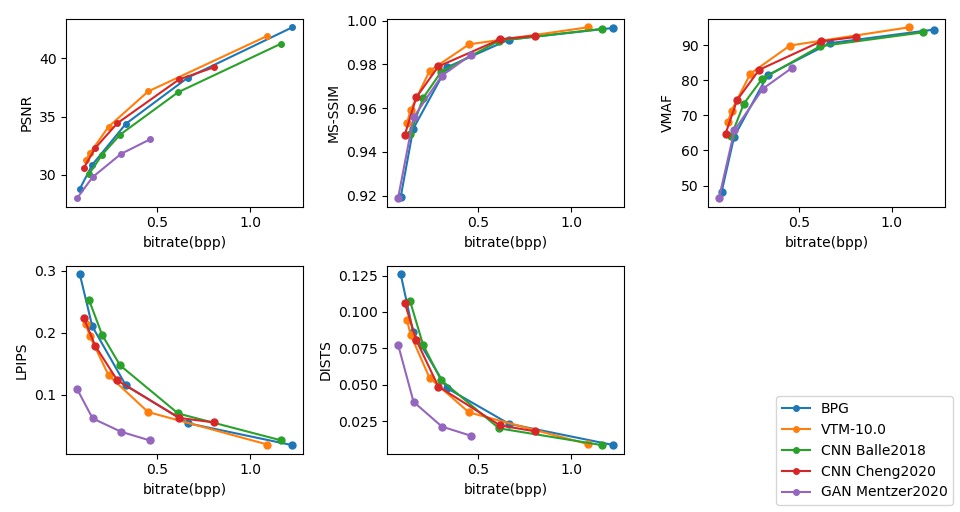}
    \caption{The average results of five metrics on all images in the image set. For the PSNR, MS-SSIM and VMAF, higher is better. For the LPIPS and DISTS, lower is better.}
    \label{fig_all}
\end{figure}

Fig.~\ref{fig_all} shows the average results of all images measured by aforementioned quality metrics. Compared with three learning-based methods, we can see that VTM performs best in PSNR, MS-SSIM and VMAF. The CNN method proposed in 2020 has close performance to VTM. The CNN method proposed in 2018 has similar performance to BPG, which are in line with our expectation. What’s interesting is the GAN method. Except for LPIPS and DISTS, the GAN method performs bad in other three evaluation metrics. However, as mentioned above, the GAN has excellent visual performance in low bit rate in our experiment, at least not worse than the two CNN methods, which means PSNR, MS-SSIM and VMAF can not assess images generated by GAN exactly. DISTS and LPIPS can distinguish the images coding by the GAN from those coding by other four methods. LPIPS learns the perceptual similarity between two images by measuring the distance of corresponding feature maps using normalized l2 distance\cite{lpips}. Different from LPIPS, DISTS learns image quality by synthesizing the texture similarity and the structure similarity of corresponding feature maps, which is detailed in \cite{DISTS}. Because both LPIPS and DISTS measures image quality on deep feature maps, we suppose that measureing images on deep features may obtain results closer to human visual perception, but it needs more evidence to verify.

\section{Conclusion}
In the paper, we describe the subjective experiment organized to compare the performance of learning-based image coding methods and traditional image coding methods with only decompressed images available. The result shows that the well-trained GAN and CNN coding methods have the potential to provide decompressed images with better visual quality than traditional coding methods in low bit-rate but have no superiority in high bit-rate. Besides, current widely used objective quality assessment methods (e.g. PSNR, MS-SSIM) can not evaluate GAN generating images well. Currently, many learning-based methods tarin one model for some specific target quality, which is inconvenient when various target quality points are required. Thus, it is significant to do some work on the method that has large quality range in one model.
%
%
%
\bibliographystyle{splncs04}
\bibliography{reference}{}
%




\end{document}